\documentclass[letterpaper, 10 pt, conference]{ieeeconf}  % Comment this line out if you need a4paper

\IEEEoverridecommandlockouts                              % This command is only needed if 
\usepackage{times}
\usepackage{graphicx}

\usepackage{subcaption}  % For subfigures

\usepackage{multicol}
\usepackage{colortbl}
\usepackage{booktabs}
\usepackage{comment}
\usepackage{times}
\usepackage{epsfig}
\usepackage{lipsum}
\usepackage{amsmath}
\usepackage{pifont}
\usepackage{amssymb}
\usepackage{arydshln}
\usepackage{color}
\usepackage[font=small, labelfont=bf]{caption}
\usepackage{url}
\usepackage{here}
\usepackage{balance}
\usepackage{xspace}
\usepackage{multirow}
\usepackage{bbding}
\usepackage[ruled,vlined]{algorithm2e}
\usepackage{stfloats}
\usepackage{wrapfig}
\usepackage{placeins}
\usepackage{makecell}
\usepackage[table,dvipsnames]{xcolor}
\usepackage{xcolor}
\usepackage{fontawesome5}
\usepackage[bookmarks=true, colorlinks=true, linkcolor=orange, citecolor=orange, urlcolor=orange]{hyperref}

\makeatletter
    \let\NAT@parse\undefined
\makeatother
\usepackage[square, numbers, sort&compress]{natbib}

\newcommand{\ours}[0]{\rowcolor{orange!10}}
\renewcommand{\baselinestretch}{1.0} %0.985

\newcommand{\MethodName}{OmniVLA}

\title{\Large \bf
\MethodName: An Omni-Modal Vision-Language-Action Model for Robot Navigation}
\author{Noriaki Hirose$^{1, 2}$, Catherine Glossop$^{1}$, Dhruv Shah$^{3}$ and Sergey Levine$^{1}$% <-this % stops a space
\thanks{$^{1}$UC Berkeley, $^{2}$Toyota Motor North America, $^{3}$Princeton university}%
}

\begin{document}
\makeatletter
\let\@oldmaketitle\@maketitle%
\renewcommand{\@maketitle}{\@oldmaketitle%
    \centering
    \vspace{3mm}
    \includegraphics[width=0.99\linewidth]{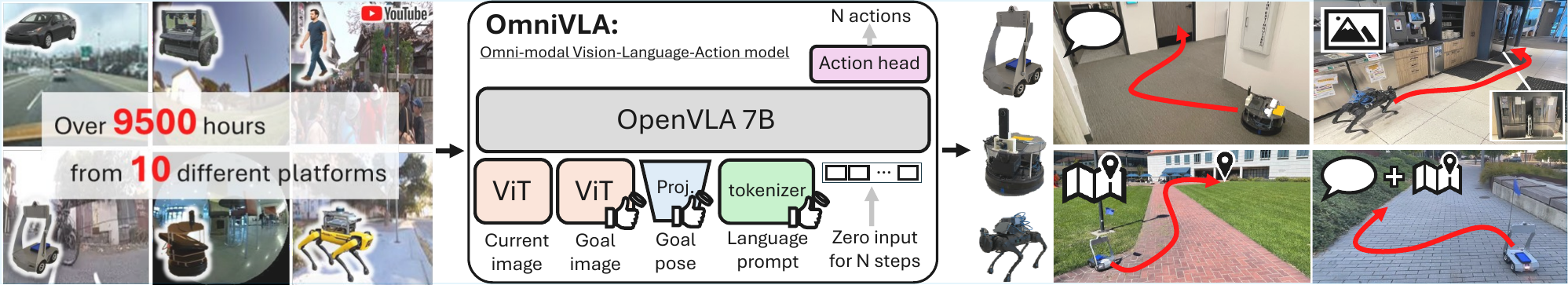}
    \captionof{figure}{We train a highly generalizable vision-based navigation policy with flexible  conditioning, leveraging over 9,500 hours of data collected across 10 different platforms. Our policy supports diverse goal modalities, including language prompts, goal poses, goal images, and their combinations, and can control a variety of robot platforms.}
    %By initializing from large pretrained visual-language-action models, our approach achieves strong out-of-distribution performance beyond the training dataset.}
    %\vspace{-1mm}
    \label{fig:pull}
}
\makeatother
\maketitle
\addtocounter{figure}{-1}
\thispagestyle{empty}
\pagestyle{empty}
%%%%%%%%%%%%%%%%%%%%%%%%%%%%%%%%%%%%%%%%%%%%%%%%%%%%%%%%%%%%%%%%%%%%%%%%%%%%%%%%
\begin{abstract}

Humans can flexibly interpret and compose different goal specifications, such as language instructions, spatial coordinates, or visual references, when navigating to a destination. In contrast, most existing robotic navigation policies are trained on a single modality, limiting their adaptability to real-world scenarios where different forms of goal specification are natural and complementary. In this work, we present a training framework for robotic foundation models that enables omni-modal goal conditioning for vision-based navigation. Our approach leverages a high-capacity vision-language-action (VLA) backbone and trains with three primary goal modalities: 2D poses, egocentric images, and natural language, as well as their combinations, through a randomized modality fusion strategy. This design not only expands the pool of usable datasets but also encourages the policy to develop richer geometric, semantic, and visual representations. The resulting model, \MethodName{}, achieves strong generalization to unseen environments, robustness to scarce modalities, and the ability to follow novel natural language instructions. We demonstrate that \MethodName{} outperforms specialist baselines across modalities and offers a flexible foundation for fine-tuning to new modalities and tasks. We believe \MethodName{} provides a step toward broadly generalizable and flexible navigation policies, and a scalable path for building omni-modal robotic foundation models.
We present videos showcasing \MethodName{}’s performance and will release its checkpoints and training code on our project page\footnote{\bf \href{https://omnivla-nav.github.io/}{\texttt{omnivla-nav.github.io}}}.
%%SL.9.6: This abstract reads very well!

\end{abstract}

%%%%%%%%%%%%%%%%%%%%%%%%%%%%%%%%%%%%%%%%%%%%%%%%%%%%%%%%%%%%%%%%%%%%%%%%%%%%%%%%
\section{Introduction}
%CG.09.02
When navigating in an environment, humans naturally use various modalities of information to infer and discover paths to goals (e.g., looking up a GPS location in a map, visual landmarks, and following language directions). However, prior work in robot navigation typically trains policies with single modalities based on narrow applications. When goals are nearby, it is convenient to describe them in a language (for example, ``move along the building and go to the entrance'', while further goals can be described more effectively as GPS coordinates. However, a truly generalist navigation policy must be able to perform tasks that require us to leverage multiple sources of information to be successful. For example, it might be most relevant to specify a combination of GPS coordinates and landmark images for tasks like autonomous delivery or inspection. While these modalities overlap significantly, they provide complementary information about the task and how it should be performed, particularly in the context of the partial view of the world that the robot receives from its sensors. Motivated by this need, this study aims to train a generalizable policy capable of navigating with goal specifications expressed in multiple modalities. By training on omni-modal goals, we aim to enable stronger and more flexible policies, ultimately acquiring a foundation model that exhibits high adaptability to novel modalities and unseen environments. 
%%SL.9.6: The idea in this paragraph is good, but the paragraph could be polished a bit to progress more logically. Right now it kind of jumps between the "human" motivation and some practical motivations. Maybe the way I would recommend to do it is to first bullet point the main ideas, then structure them into an argument where each idea is presented in turn.

Training foundation models requires that we leverage as much data as possible. With recent data collection efforts in the robotics community, it has become feasible to train powerful generalist navigation policies. However, these policies are typically trained with only a single kind of task representation, such as egocentric images, 2D poses, or natural language. This limits the datasets that can be used to those that accord with the desired task representation (e.g., only datasets with language labels), restricts how the model can be used at test time, and potentially limits how the model processes the task and observation -- i.e., more spatial or visual tasks might improve the model's spatial reasoning abilities, while language tasks might provide a complementary benefit in enhancing understanding of semantics, much like task mixtures for multi-modal language models~\cite{lin2024vila,deitke2025molmo}.
%%SL.9.6: I rewrote the above a bit, the last point is probably the most interesting (I think this is also the point you were trying to make before), but since it's a bit speculative, it would help to bolster it by citing some VLM paper that talks about task mixtures. Could just cite a couple recent VLM papers, like VILA and Molmo or something

In this study, we propose a family of {\bf Omni}-Modal {\bf V}ision-{\bf L}anguage-{\bf A}ction Models ({\bf \MethodName{}}) for autonomous navigation that can ingest goals expressed in multiple modalities, leveraging information across modalities, and achieving a more flexible navigation policy. We train our model with goals specified through three primary modalities: (1) 2D poses, (2) egocentric images, and (3) natural language. By simultaneously learning to interpret these different modalities, the model must develop a richer understanding of the geometric, visual, and semantic information of the task, resulting in a more powerful navigation model. 
Moreover, our method allows the user to instruct the robot with multiple modalities, making it more user friendly and directly allowing the policy to leverage more than one kind of information about a goal. For example, a user can specify a target pose and provide instructions on \emph{how} to reach it through language. 

To train these policies, we compose several design choices into one system, resulting in a flexible and general navigation policy. We use an expressive vision-language-action (VLA) model as the base model~\cite{kim24openvla}, enabling us to leverage internet-scale knowledge from the VLM backbone~\cite{karamcheti2024prismatic} and the representations learned during fine-tuning on cross-embodiment robot data~\cite{o2024open}. As a result, our policy exhibits strong generalization and fine-tuning capabilities, following language instructions not seen in the training data, and adapting to completely new modalities. Additionally, we address the problem of modality imbalance and scarcity by using modality dropout during training, and modality masking during inference. This ensures that our policy attends to all available goal modalities and learn from cross-modal goal representations
% the modalities and their combinations 
across all datasets. 

% The main contribution of our work is \MethodName, a foundational navigation VLA model that demonstrates strong performance under omni-modal conditioning. While the building blocks of \MethodName{} draw heavily on prior works, it is 
OmniVLA is the first end-to-end VLA model for navigation that unifies diverse task modalities. Our results demonstrate strong performance across all modalities, surpassing \emph{specialist} baselines trained for a single task or goal modality. 
Furthermore, we demonstrate that OmniVLA can be efficiently fine-tuned to new goal modalities and new environments using a limited dataset. %Furthermore, we demonstrate that OmniVLA can be efficiently fine-tuned to new goal modalities using as little as 1 hour of labeled data for a new modality. 
We believe \MethodName{} will serve as a valuable resource for future navigation research, both as a recipe for a generalist model and as a pre-trained checkpoint for fine-tuning to specific modalities.
% We may want to add specifics here or in the abstract for comparisons to single modality polices / multiple inputs

\section{Related work}
In navigation tasks, we can specify goals with various modalities, such as egocentric images~\cite{shah2023gnm,hirose2022exaug,shah2023vint,sridhar2024nomad}, 2D poses~\cite{shah2022viking,hirose2025learning}, and language~\cite{gadre2023cows,hirose2024lelan,glossop2025cast,cheng2024navila}. 
For egocentric image-conditioned navigation, prior works~\cite{shah2023gnm,hirose2022exaug,shah2023vint,sridhar2024nomad} combine publicly available datasets across various robot embodiments to train generalist policies. These policies work well in indoor environments, which require rich visual information and do not allow for reliable access to GPS.
2D pose-conditioned policies are most successful in long-horizon tasks in outdoor environments with GPS localization~\cite{shah2022viking}. MBRA\cite{hirose2025learning} introduces a model-based reannotation approach to leverage large-scale data sources to perform more challenging long-distance navigation tasks conditioned on 2D goal poses. 

Language-conditioned navigation offers a user-friendly, flexible interface to instruct robots to reach a specific target object or area in a given environment, even over long distances~\cite{gu2022vision,gadre2023cows,shah2023lm}. However, robust instruction-following demands language understanding beyond simple object references. Early works relied on pre-trained language encoders~\cite{mees2022matters,gadre2023cows}, while recent approaches use powerful VLM backbones, either using them directly to perform navigation tasks~\cite{sathyamoorthy2024convoicontextawarenavigationusing} or fine-tuning them on robot data~\cite{xu2024mobility}. Other works have extended to using counterfactual action generation~\cite{glossop2025cast} and non-robot data~\cite{cheng2024navila} to improve training of these models. LeLaN~\cite{hirose2024lelan} leverages both robot and non-robot data to learn a generalized language-conditioned navigation policy, using a model-based approach to generate counterfactual actions that reach target objects along with language prompts derived from VLM reasoning. While such synthetic action commands and language prompts enable scalable training~\cite{glossop2025cast,cheng2024navila,hirose2024lelan}, their inaccuracy can become a performance bottleneck.

In manipulation, recent robotic foundation models (RFMs) aim to unify vision, language, and action for generalization~\cite{reed2022generalist,driess2023palm}. Evolving from early multi-modal transformers to large-scale systems focused on dexterity and real-world deployment, they mark a shift toward scalable frameworks capable of robust, versatile robotic control~\cite{o2024open,team2024octo,black2024pi_0}. Some manipulation works leverage large datasets with different observation and action spaces as well as conditioning (language, poses, etc.) by masking unavailable inputs during training~\cite{lynch2019play,team2024octo,doshi2024scaling} and demonstrate the benefits of simultaneously training on multiple input types~\cite{Myers23-grif}.

Motivated by the success of previous work in the manipulation domain, \MethodName{} goes beyond prior navigation work to leverage as much robot and non-robot data as possible across multiple modalities and embodiments, learning a policy that can condition on goal images, language, and GPS. Notably, \MethodName{} is trained on almost 10,000 hours of real-world navigation data, which is the largest pre-training dataset for an end-to-end navigation policy to our knowledge.

\begin{figure}[t]
  \vspace{1mm}
  \begin{center}
      \includegraphics[width=1.0\hsize]{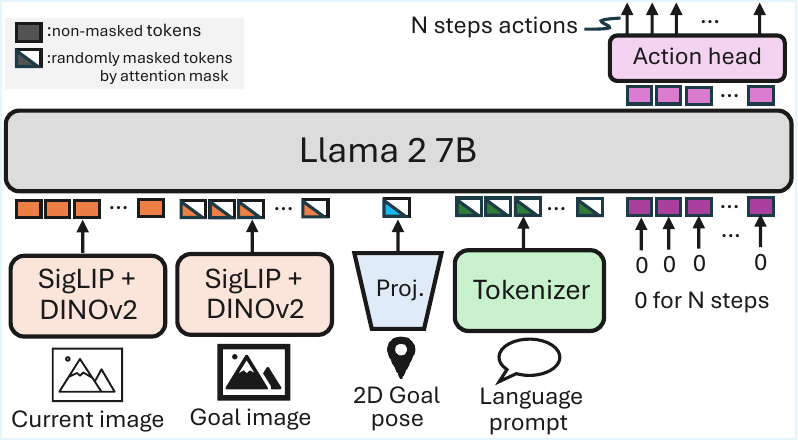}
  \end{center}
    \caption{\small {\bf Network architectures for multi-modal vision-based navigation.} Our design builds on existing large VLA checkpoints, adding a visual backbone and a projector to condition on egocentric goal images and 2D goal poses. During training, we randomly mask tokens for these modalities and the language prompt.}
    %%SL.9.6: it would be good to expand this caption to better explain the architecture
  \label{f:network}
  \vspace*{-1mm}
\end{figure}
\begin{table*}[t]
  \vspace{2mm}
  \begin{center}
  %\vspace{-2mm}
  \resizebox{2.02\columnwidth}{!}{
  \begin{tabular}{llcccccccccc} \toprule 
    \multicolumn{4}{c}{\textbf{Dataset}} & \multicolumn{3}{c}{\textbf{Action}} & \multicolumn{4}{c}{\textbf{Modality}} & \multirow{2}{*}{\textbf{Environment}} \\
    \cmidrule(lr){1-4} \cmidrule(lr){5-7} \cmidrule(lr){8-11}     
    Name & Platform & Max. speed & Used Hrs & Raw & Hz & Labels & Lang. & Pose & Ego. & Sate. & \\ \midrule
    GNM mixture~\cite{shah2023gnm} & 6 platforms & 0.5 -- 5.0 m/s & 62.0 & \checkmark & 3.0 & raw & & \checkmark & \checkmark & & Off-road, office, sidewalks \\    
    LeLaN mixture~\cite{hirose2024lelan} & 3 platforms & 0.5 -- 1.0 m/s & 128.7 & & 3.0 & NoMaD & \checkmark$\star$ & \checkmark$\star$ & & & Office, home, sidewalks \\
    Frodobots-2K~\cite{frodobots2k} & ERZ & 1.0 m/s & 700 & \checkmark & 3.0 & MBRA & & \checkmark & \checkmark & \checkmark & Sidewalks \\    
    BDD-V~\cite{xu2017end} & Car & 20.0 m/s & 8680 & \checkmark & 1.0 & \hspace{2mm}MBRA $\diamond$ & & \checkmark & \checkmark & & Road  \\             
    \bottomrule       
  \end{tabular}%
  }
  \end{center}
  %\vspace{-2mm}
  \caption{{\bf Training dataset mixture.} Our training mixture consists of 9,500 hours across 10 different platforms, including human-collected data, and covers a diverse set of environments. GNM and LeLaN are themselves mixtures of 7 and 5 (respectively) publicly available datasets. The details are shown in the Appendix \hyperref[app:gnm_lelan]{D}. We use the satellite modality in the Frodobots dataset for evaluation of \MethodName's ability to adapt to a new task. $\star$ indicates labels that are synthetically generated, and $\diamond$ indicates that a customized version of MBRA was used to generate the refined action labels, described in the Appendix~\hyperref[app:bdd_annot]{B}.} %\hyperref[sec:reannot]{Section \ref{sec:reannot}}
  \label{tab:dataset}    
  %\vspace{-5mm}
\end{table*}
\section{An Omni-Modal VLA for Navigation}
%
%In this paper, 
We present \MethodName, an end-to-end navigation policy with omni-modal conditioning. \MethodName{} effectively captures core navigation capabilities, such as collision avoidance and path following, and can perform complex goal-conditioned navigation behaviors through omni-modal user inputs. 
Furthermore, to train \MethodName{}, we leverage numerous public datasets—each containing at least one modality—resulting in a model with strong generalization and improved adaptability to new modalities and environments.
%%SL.9.6: Perhaps we can use a different term than "flexible training method", to avoid a misunderstanding where readers think this is an algorithm paper
%
\subsection{\MethodName{} Architecture}
\label{sec:net}

We build \MethodName{} on top of a high-capacity VLA architecture, which contains knowledge from Internet-scale pre-training (for the base VLM) and cross-embodiment robot actions (for the base VLA). We modify the architecture to enable flexible goal-conditioning and omni-modal representation learning, while preserving the strong vision-language priors in the base model.

Fig.~\ref{f:network} illustrates the network architecture, built on top of OpenVLA~\cite{kim24openvla}, a 7B-parameter VLA model. We process the robot's current visual observations using a visual encoder. For goal-conditioning, our architecture supports three different modalities --- visual, positional, and language --- which can also be specified simultaneously. We project each individual goal modality into a shared token space, which serves as the input of the LLM backbone. We use ``modality dropout'' to flexibly mask the goal modalities during training and inference. For the action output, we follow OpenVLA-OFT~\cite{kim2025fine} and add a linear action head to the LLM output to generate a sequence of actions $\{ \hat{a}_i \}_{i=1 \ldots N}$.

We also implement a smaller ``edge'' version of our \MethodName{} architecture built on top of ViNT~\cite{shah2023vint}, a 50M-parameter navigation transformer. We modify this to add multiple goal encoders and perform modality dropout in the same way as our base model (see Appendix~\hyperref[app:vint]{A}). In real-world evaluation~\ref{sec:evaluation_setup}, we find \MethodName{}-edge to be a very compelling choice for resource-constrained deployment, where inference of large VLA models is not feasible.

% \subsection{Randomized multi-modal fusion learning}
% \label{sec:randfusion}
%
\subsection{Training \MethodName{}}
While using multi-modal inputs is enticing, training policies to accept omni-modal inputs requires compiling robot datasets that support training and addressing the relative imbalance and scarcity of the available modalities.
%While using multi-modal inputs is enticing, training policies to accept omni-modal inputs requires that we compile robot datasets that support training them and address the relative imbalance and scarcity of the various modalities available.

\textbf{Training data.} Table~\ref{tab:dataset} lists the publicly available datasets for vision-based navigation. This data is comprised of 13 publicly available datasets and contains 9,500 hours across 10 different embodiments. While most of these datasets include manually collected labels, we also include data that has synthetic labels, which may be noisier, and actions refined with a re-annotation process. To our knowledge, we compile the largest mixture of navigation datasets with highly diverse environments, embodiments, and modalities to train \MethodName{}. 

While large datasets enable generalization, large-scale data collection efforts can result in more noise and therefore, be less accurate. Following \cite{hirose2025learning} and \cite{hirose2024lelan}, we use synthetic actions generated with MBRA~\cite{hirose2025learning} for the Frodobots dataset and NoMaD~\cite{sridhar2024nomad} for the LeLaN dataset during training. Since existing reannotation approaches cannot account for the large embodiment gap of the BDD-V~\cite{xu2017end} dataset (an autonomous vehicle dataset vs. the small robot datasets we use otherwise), we train a reannotation model to generate reasonable synthetic actions, making it possible to use this data for training, similar to \cite{hirose2025learning}. However, we found it was reasonable to train directly with the high-quality raw actions available in the GNM mixture. The details about the BDD-V~\cite{xu2017end} dataset are shown in the Appendix~\hyperref[app:bdd_annot]{B}.

\textbf{Training method.} Following prior work for manipulation VLAs~\cite{team2024octo}, we use a form of random dropout to train on all available modalities, resulting in a more efficient and generalized model. To train \MethodName{}, we construct a batch of samples from the datasets included in Table~\ref{tab:dataset}. For each sample, we then independently sample from the available goal modalities to form the conditioning input, which we call $t_m$. For example, for the GNM mixture, the conditioning input can be chosen from 2D pose, egocentric goal image, or their combination. Naturally, we get coverage over all modalities and datasets while using this dropout mechanism to improve training stability.

For each training step, we construct an attention mask that excludes unused or unavailable modalities (filled with random values) so that only the selected modalities are attended to. Training on these mixed-modality batches encourages the model to better represent goal information, yielding improved representations for generalization and fine-tuning.
%For each training step, we construct an attention mask that excludes unused or unavailable modalities (which we fill with random values) so that only the selected modalities are attended to. We find that by training on these mixed modality batches, the model must learn to better represent the goal information, resulting in what we believe to be better downstream representations for generalized performance and fine-tuning. 

We train the \MethodName{} policy described by $\{ \hat{a}_i \}_{i=1 \ldots N} = \pi_{\theta}(I_c, I_g, p_g, l_g, t_m)$, where $I_c$, $I_g$, $p_g$, $l_g$ and $t_m$ are the egocentric current image, the egocentric goal image, the 2D goal pose, the language prompt and the randomly selected modality, respectively. We calculate the objective $J$ with $J_{il} = \frac{1}{N}\sum_{i=1}^{N}(a^{ref}_i - \hat{a}_i)^2$ to imitate the $N$-step action reference $\{ a^{ref}_i \}_{i=1\ldots N}$ and update $\theta$ to minimize $J$. 
Following \cite{hirose2024lelan}, the examples in the LeLaN dataset use an additional task-specific objective to encourage object reaching behavior, where the final action is trained to be close to the target object. 
% This objective was used with LeLaN in \cite{hirose2024lelan}, and we directly incorporate it in our model as well.
%Following \cite{hirose2024lelan}, we introduce additional task-specific objectives to encourage object reaching behavior.
%%SL.9.13: if this is only on LeLaN, maybe say:
%Following \cite{hirose2024lelan}, the examples in the LeLaN dataset use an additional task-specific objective to encourage object reaching behavior, where the final action is trained to be close to the target object. This objective was used with LeLaN in \cite{hirose2024lelan}, and we directly incorporate it in our model as well.
Details of the objectives are shown in the Appendix~\hyperref[app:obj]{E}.

\textbf{Training details.} We set the action chunk size $N$ to 8 at 3 Hz, corresponding to 2.4 seconds for all models. For each batch, we sample the data at a ratio of LeLaN : GNM : Frodobots : BDD-V = 4:1:1:1 to balance modalities. Since we cannot secure a sufficiently large batch size for some models even on a server with multiple GPUs, we accumulate the gradient for several steps to stabilize the training process. 
%We design the step number for gradient accumulation to effectively achieve an equivalent batch size to be over 200. 
%We train on a machine with eight H100 GPUs for the \MethodName{} initializing with the OpenVLA checkpoints. 
In training \MethodName{} with OpenVLA checkpoints on eight H100 GPUs, we use a per-GPU batch size of 7 and accumulate gradients for 4 steps, yielding an effective batch size of 224 (=7×8×4). In addition, we apply LoRA to limit the learnable parameters to about 5 $\%$, allowing us to maximize the batch size and balance training speed with training stability. Note that LoRA is only used for the OpenVLA-based model due to its large model size. The other training settings, such as learning rate, language tokenization, normalization, and so on, are the same as the default setting in the original code for each model type.
\section{Experimental Setup}
\label{sec:evaluation_setup}
We begin by describing our setup for evaluating omni-modal navigation on our real-world robot platforms. We conduct extensive real-world evaluations and compare against state-of-the-art specialist and generalist baselines.

\subsection{Navigation tasks}
%
% We first describe navigation tasks for each modality, the baselines, as well as the robots used in our experiments.
We consider the following three navigation tasks:

\textit{{\bf Language-conditioned navigation:}} 
We evaluated \MethodName{} on language prompts that not only direct the robot toward the target location but also specify how it should behave along the way. We conducted evaluations in 40 environments, including an office, a kitchen area, an entrance hall, a public park, and sidewalks, using diverse language prompts. The goals were placed 5–30 meters from the robot’s initial position. To assess the benefit of large pre-trained models, we introduced out-of-distribution (OOD) language prompts that go beyond the instructions present in the training data. While the training data primarily contained object-reaching instructions such as ‘move toward X’, we designed OOD prompts that specified how to navigate toward the target in half of the trials. Following CAST~\cite{glossop2025cast}, we crafted these prompts and evaluated the robot’s behavior based on its adherence to instructions. We also selected 17 environments where obstacles were placed between the robot’s start and the target, making the tasks more challenging and testing the core navigation abilities of the policy.

\textit{{\bf Egocentric goal image-conditioned navigation:}} 
With egocentric goal images, our policy is tasked with navigating the robot to target locations up to 3 meters away. To extend this range, we follow prior vision-based approaches~\cite{shah2023vint,savinov2018semi,hirose2019deep} and employ topological memory for navigation in 8 different environments, enabling navigation to more distant goals. To build the goal graph, we record image observations at 1 Hz. During deployment, we initialize from the first observation and, at each time step, estimate the closest node as the current location, as in \cite{shah2023vint,sridhar2024nomad}. The image from the next node is then provided as the goal image $I_g$ to our policy.

\textit{{\bf Goal pose-conditioned navigation:}} 
When conditioned on 2D goal poses, our policy navigates to targets 25–100 meters from the initial robot position. We use GPS to estimate the robot's location and the location of the goal. At each step, we compute the pose of the robot relative to the target pose, $p_g$. For evaluation, we selected 7 environments and ran 3 trials at different times in each, accounting for GPS jitter.
%When conditioned on 2D goal poses, our policy can navigate to targets located 25–100 meters from the initial robot position. We use GPS to estimate the robot’s position and define the goals. At each step, we compute the relative goal pose $p_g$ toward the next target. For evaluation, we selected 7 environments and conducted 3 trials at different times in each, accounting for GPS jitter.
%
\subsection{Mobile Robot Platforms}
\begin{figure}[t]
  \vspace{2mm}
  \begin{center}
      \includegraphics[width=0.99\hsize]{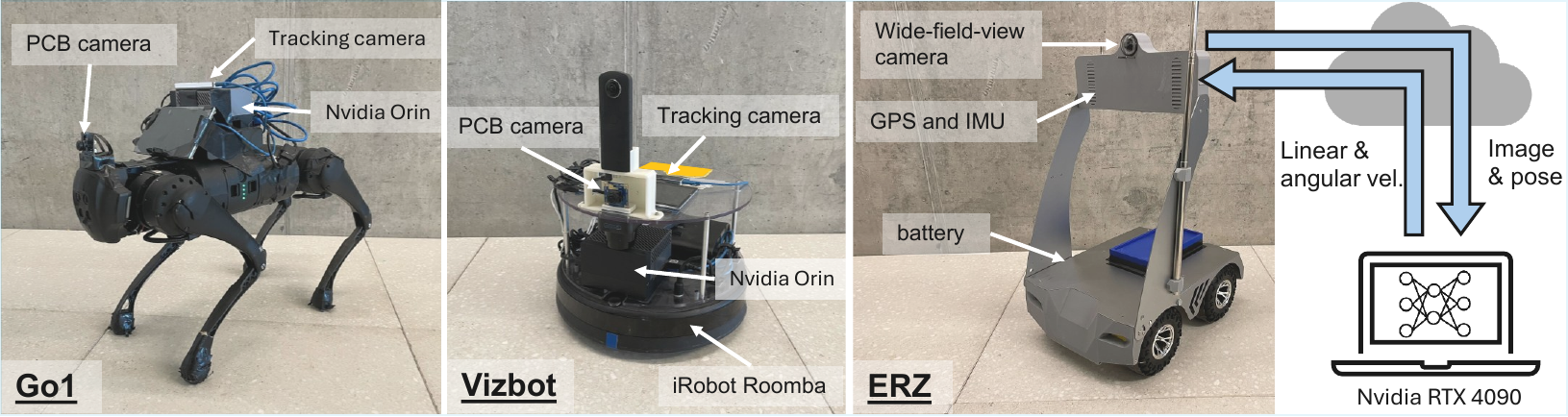}
  \end{center}
  %\vspace{2mm}
  \caption{\small {\bf Overview of the robotic platforms in our evaluation.} Our local PC with an NVIDIA RTX 4090 receives front-camera images and pose signals, and sends velocity commands from our policy to control the robot over the internet.}
  \label{f:robots}
  %\vspace{-2mm}
 \end{figure}

We evaluate \MethodName{} on the FrodoBots ERZ platform~\cite{erz}, a low-cost mobile robot in Fig.~\ref{f:robots}. It is equipped with multiple sensors, including front and rear cameras, GPS, an IMU unit (gyroscope, accelerometer, and compass), and wheel velocity sensors on all four wheels. All sensor streams are accessible through a web API, and our trained navigation policies interface with the robot by sending linear and angular velocity commands.

To assess cross-embodiment generalization, we evaluate our policies on two additional platforms: the VizBot~\cite{niwa2022spatio} wheeled robot and the Unitree Go1~\cite{go1} quadruped.
\subsection{Baselines}
In our evaluation, we compare \MethodName{} against seven baselines across all modalities. This includes model-free navigation policies trained from scratch~\cite{shah2023vint, sridhar2024nomad, hirose2025learning}, methods that can leverage internet-scale videos~\cite{hirose2024lelan}, methods using off-the-shelf visual representations like CLIP~\cite{gadre2023cows}, as well as state-of-the-art VLA models~\cite{glossop2025cast}. For LeLaN, CounterfactualVLA, ViNT, MBRA-pose, and MBRA-image, we use the authors' original implementation and checkpoints for evaluation. Below, we describe two baselines that differ slightly from their original implementations.

{{\bf NoMaD~\cite{sridhar2024nomad}:}} For 2D goal pose-conditioned navigation, we run the NoMaD policy in exploration mode to generate 30 candidate trajectories. We then select the trajectory whose final predicted position is closest to the target pose $p_g$ and use it to control the robot. For egocentric goal image-conditioned navigation, we follow the original NoMaD implementation using the goal image $I_g$.

{{\bf CoW~\cite{shah2023vint}:}} For language-conditioned navigation, we provide the current observation and prompts describing the target object to the OWL-ViT B/32 detector~\cite{minderer2022simple}, reported as the strongest model in the original paper, to estimate the object’s bounding box. Following \cite{hirose2024lelan}, we crop the point cloud within the box and compute the median point as the target object pose. To ensure fair comparison with our approach, which relies solely on a single RGB camera without depth or LiDAR, we estimate depth using Depth360~\cite{hirose2022depth360} and project it to reconstruct the point cloud. A state lattice motion planner is then used to generate velocity commands.

\textbf{Other VLA backbones:} To further understand the role of VLA architectures and pre-training, we also implement our omni-modal goal-conditioning strategy for the 1B MiniVLA~\cite{belkhale2024minivla} and the 500M SmolVLA~\cite{shukor2025smolvla}. Please see the Appendix~\hyperref[app:pretrained_vla]{C} for details on these architectures.
\begin{table}[t]
  \vspace{2mm}
  \begin{center}
  %\vspace{-2mm}
  \resizebox{1.0\columnwidth}{!}{
  \begin{tabular}{lc@{\hskip 6pt}c@{\hskip 6pt}c@{\hskip 6pt}cc@{\hskip 6pt}cc@{\hskip 6pt}c} \toprule
    \textbf{Method} & \multicolumn{4}{c}{\textbf{Language}} & \multicolumn{2}{c}{\textbf{2D Pose}} & \multicolumn{2}{c}{\textbf{Image}} \\
    \cmidrule(lr){2-5} \cmidrule(lr){6-7} \cmidrule(lr){8-9}
    & SR & Behavior & SR$^S$ & SR$^C$ & SR & Prog. & SR & Prog. \\ \midrule
    CoW~\cite{gadre2023cows} & 0.30 & 0.05 & 0.55 & 0.24 & -- & -- & -- & -- \\
    LeLaN~\cite{hirose2024lelan} & 0.43 & 0.15 & 0.64 & 0.35 & -- & -- & -- & -- \\
    CounterfactualVLA~\cite{glossop2025cast} & 0.33 & 0.45 & 0.18 & 0.24 & -- & -- & -- & -- \\
    MBRA-pose~\cite{hirose2025learning} & -- & -- & -- & -- & 0.86 & 0.92 & -- & -- \\
    NoMaD~\cite{sridhar2024nomad} & -- & -- & -- & -- & 0.33 & 0.47 & 0.63 & 0.77 \\
    ViNT~\cite{shah2023vint} & -- & -- & -- & -- & -- & -- & 0.50 & 0.68 \\
    MBRA-image~\cite{hirose2025learning} & -- & -- & -- & -- & -- & -- & \bf{1.00} & \bf{1.00} \\
    SmolVLA~\cite{shukor2025smolvla} & 0.10 & 0.15 & 0.00 & 0.12 & 0.38 & 0.70 & 0.38 & 0.45 \\
    MiniVLA~\cite{belkhale2024minivla} & 0.23 & 0.15 & 0.27 & 0.12 & 0.43 & 0.75 & 0.25 & 0.36 \\
    \ours \MethodName{}-edge & 0.60 & 0.25 & 0.82 & 0.47 & 0.91 & 0.95 & \bf{1.00} & \bf{1.00} \\
    \ours \MethodName{} & \bf{0.73} & \bf{0.65} & \bf{1.00} & \bf{0.65} & \bf{0.95} & \bf{0.98} & \bf{1.00} & \bf{1.00} \\
    \bottomrule
  \end{tabular}%
  }
  \end{center}
  %\vspace{-2mm}
  \caption{{\bf Quantitative analysis for conditioning on single modalities.} SR and Prog. indicate the success rate and the partial progress towards the goal, respectively. ``SR$^S$'' averages over simple experiments without obstacles. ``SR$^C$'' averages over complex experiments with obstacles in the environment. Behavior indicates the success rate in following OOD language prompts.}
  %\vspace{-3mm}
  \label{tab:individual}
\end{table}
\begin{table}[t]
  %\vspace{-2mm}
  \begin{center}
  \resizebox{1.0\columnwidth}{!}{
  \begin{tabular}{lc@{\hskip 3pt}c@{\hskip 3pt}c@{\hskip 3pt}cc@{\hskip 6pt}c@{\hskip 6pt}c@{\hskip 6pt}c} \toprule
    \multirow{2}{*}{\parbox{1.5cm}{\textbf{Train Modality}}} & \multicolumn{4}{c}{\textbf{Dataset}} & \multicolumn{4}{c}{\textbf{Test Modality}} \\
    \cmidrule(lr){2-5} \cmidrule(lr){6-9}
    & GNM & LeLaN & Frod. & BDD & Lang. & 2D pose & Ego. image & Sate. image \\ \midrule
    Language & & \checkmark & & & 0.43 & -- & -- & -- \\
    2D pose & \checkmark & & \checkmark & & -- & 0.86 & -- & -- \\
    Ego. image & \checkmark & \checkmark & \checkmark & & -- & -- & {\bf 1.00} & -- \\
    Sat. image & & \checkmark & & & -- & -- & -- & 0.19 \\
    \ours Omni-modal & \checkmark & \checkmark & \checkmark & \checkmark & {\bf 0.60} & {\bf 0.91} & {\bf 1.00} & \bf{0.57} \\
    \bottomrule
  \end{tabular}%
  }
  \end{center}
  %\vspace{-2mm}
  \caption{{\bf Ablation study of multi-modal training \MethodName{}-edge.} We evaluate the performance of \MethodName{}-edge when trained on single and multiple modalities and evaluated with single modality task representations.}
  \label{tab:ablation}
  %\vspace{-3mm}
\end{table}
%
%
%\newpage
\section{Evaluating Omni-Modal Navigation}
\label{sec:evaluation}

To evaluate our \MethodName{} policies, we focus our experiments on answering the following questions:
\begin{itemize}
    \setlength{\parskip}{0cm}
    \setlength{\itemsep}{0cm}
    \item[{\bf Q1}] Does omni-modal pre-training outperform single-modality navigation policies?
    \item[{\bf Q2}] Can \MethodName{} follow a composition of multiple goal modalities?
    \item[{\bf Q3}] Can \MethodName{} be adapted to new goal modalities, environments, and embodiments?
    %%SL.9.6: these questions feel a bit vague, in the sense that it's not clear if there is anything that would confirm or deny these questions, can we make them more scientifically falsifiable?
    %%NH.9.9: I fixed.
\end{itemize}
\subsection{\MethodName{} vs. single-modality policies}
For {\bf Q1}, we conduct a comparative analysis with single-modality policy baselines, with results summarized in Table~\ref{tab:individual}. The evaluation metrics are provided in the table caption. Our largest model, \MethodName{}, demonstrates a clear advantage over each modality-specific baseline. This improvement stems from our policy’s ability to learn more generalized navigation capabilities from our large, highly diverse training mixture, which is substantially larger than the datasets available for individual modalities. Additionally, we observe that the choice of pre-trained VLA architecture and pre-training data has a significant impact on the performance. Notably, \MethodName{} demonstrates strong generalization on language- and pose-conditioned tasks, outperforming the strongest single-modality specialist. 

\textbf{Architecture comparison.} While all the \MethodName{} variants demonstrate the ability to generalize across modalities, \MethodName{} outperforms the smaller VLA models by a wide margin. We notice that MiniVLA and SmolVLA perform quite poorly, which can be primarily attributed to the wide gap between their pre-training domains (MiniVLA primarily trained on manipulation data) and limited capacity (SmolVLA might not have enough capacity to learn useful cross-embodiment representations). In contrast, \MethodName{}-edge adopts a more specialized architecture for navigation tasks (described in the Appendix~\hyperref[app:vint]{A}), which enables it to achieve exceptional performance for its size. However, there remains a substantial gap between \MethodName{}-edge and \MethodName{} on language following, highlighting the benefits of the vision-language priors inherited from pre-trained VLMs.

\textbf{Image-conditioned performance.} \MethodName{} matches the performance as the strongest single modality baseline, MBRA, achieving 100\% success, and outperforming NoMaD and ViNT. While NoMaD and ViNT are trained on the more limited GNM mixture, MBRA leverages a much more diverse pre-training dataset that enables strong generalist performance. \MethodName{} is trained on a combination of these data sources, and is able to match this strong performance. 

\textbf{Pose-conditioned performance.} \MethodName{} achieves a 9\% performance improvement in success rate (binary) and partial progress towards the goal over the strongest \emph{specialist} baseline, MBRA-pose. While both policies leverage the same large datasets for pose-conditioned navigation, with a high-capacity model and additional data, \MethodName{} is able to outperform prior methods. This observation is particularly impressive for VLA-based policies, as this is not well represented in the VLM/VLA pre-training.

\textbf{Language-conditioned performance.} \MethodName{} demonstrates a large improvement in goal-reaching and language following on the diverse set of language prompts we evaluate. While both \MethodName{} and LeLaN are trained only with in-domain language prompts from the LeLaN dataset, the performance gap between \MethodName{} and LeLaN is particularly evident on “Behavior”, which measures success in following OOD language prompts. This result suggests that leveraging a larger pre-trained LLM provides stronger priors for navigation tasks. \MethodName{} also outperforms CounterfactualVLA, even though CounterfactualVLA is trained with more diverse language prompts, as \MethodName{} leverages a larger backbone and a larger training mixture, which makes it more general and flexible to OOD prompts. We also evaluated NaVILA~\cite{cheng2024navila} using our prompts in the same environment. However, NaVILA fails, scoring 0.0 on all metrics, due to a domain gap in prompt style: it requires detailed, step-by-step instructions such as “Turn right, move straight, and stop when you see X,” even when X is already visible and nearby, which is not in line with the open-set, natural language prompt format we target.

% \MethodName{}-Mid and \MethodName{}-Small fail to learn stronger robotic behaviors due to their limited capacity and the task gap between navigation and their original pre-training domains, which are manipulation and language answering. In contrast, \MethodName{}-edge adopts a different architecture specifically for navigation tasks (described in the Appendix~\hyperref[app:vint]{A}), which allows it to achieve the second-best performance. However, there remains a substantial gap between \MethodName{}-edge and \MethodName{} on “Lan. f.”, highlighting the advantage of stronger priors from large pre-trained models.
%%SL.9.7: I wonder if there is a better way to organize this, where we first address generalization, performance, OOD language, etc., and then have a separate paragraph discussing different pretrained backbones? The idea is to provide some degree of prioritization about which claims are important and which are secondary. The important claims I think are about how the method overall works well, and the secondary claims are about which specific backbone is better. Mixing these together can create confusion about which findings we consider more/less important
%%NH.9.9: I modified according to the Sergey's suggestions.

\begin{figure*}[t]
    \vspace{2mm}
    \centering
    % Left figure (1/3 width)
    \begin{minipage}[t]{0.33\textwidth}
        \centering
        \includegraphics[width=0.99\hsize]{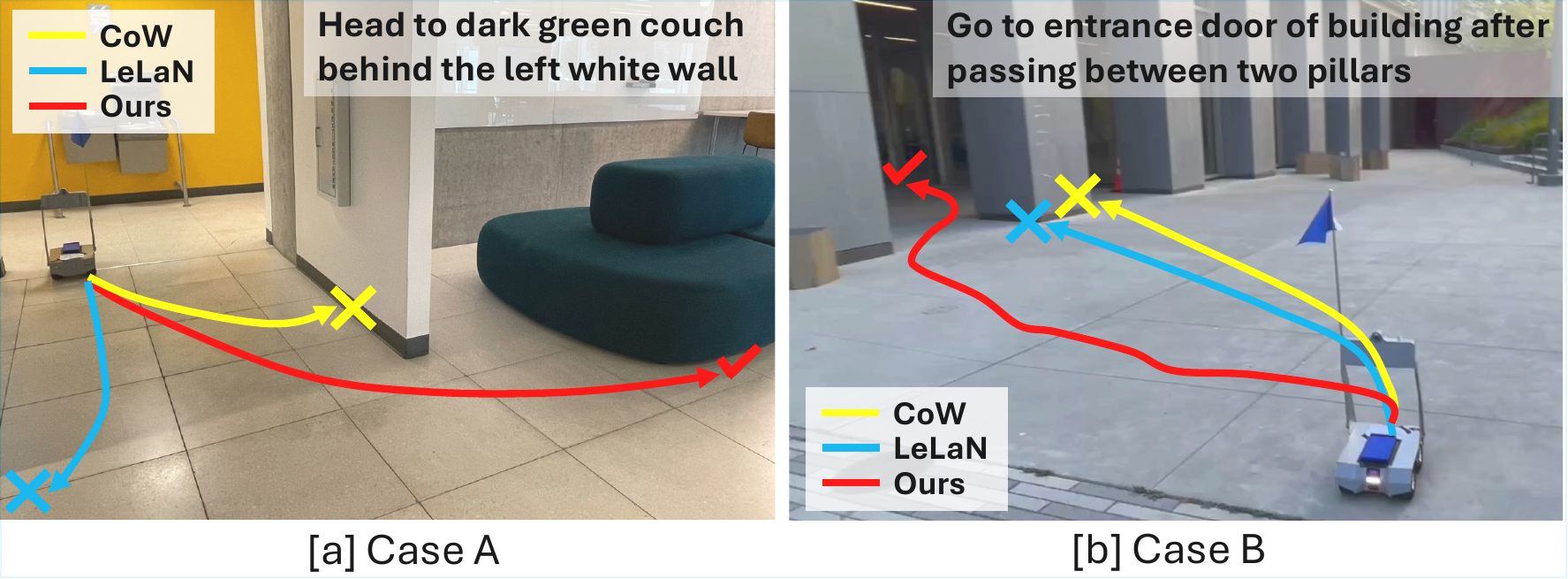}
	    \caption{\small {\bf Visualization of language-conditioned navigation rollouts.} \MethodName{} can follow OOD language instructions in various indoor and outdoor environments.}
        \label{f:vis_lan_nav}
    \end{minipage}%
    \hspace{0.005\textwidth}
    \begin{minipage}[t]{0.65\textwidth}
        \centering
        \includegraphics[width=0.99\hsize]{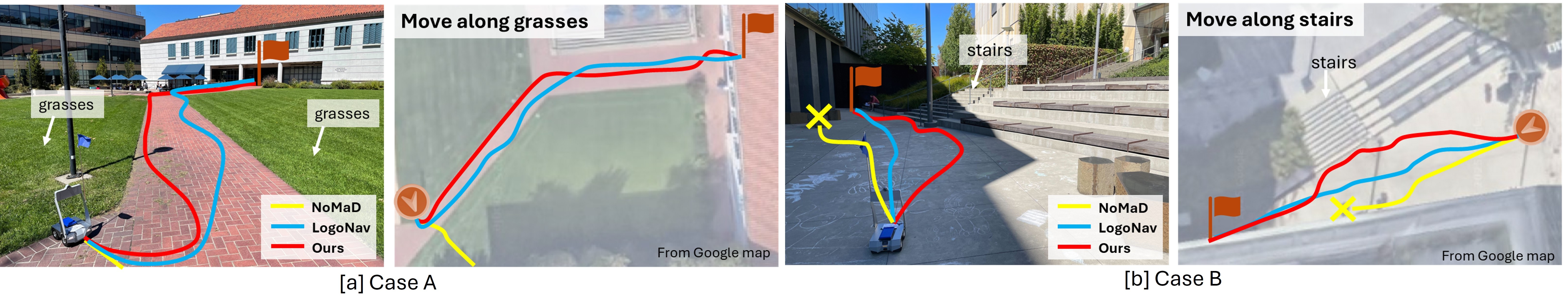}
	    \caption{\small {\bf Visualization of goal pose- and language-conditioned navigation rollouts.} Conditioned on OOD language and a goal pose, our policy can perform complex, long-horizon navigation tasks, reaching distant goals ({\color{orange}\faFlag}) while following behavioral instructions.}
        \label{f:vis_pos_lan_nav}
    \end{minipage}
    %\vspace{-2mm}    
\end{figure*}

Figure~\ref{f:vis_lan_nav} illustrates rollouts of \MethodName{} and the baselines on two language-conditioned navigation tasks. Both scenes are particularly challenging as the target objects are not visible from the robot’s initial position, requiring the policy to use the other information in the prompt to reach the goal. Our method successfully follows the language instructions and reaches the target object, whereas the baselines LeLaN and CoW fail, navigating instead toward the incorrect object.

\textbf{Dataset ablation.}
We conduct an ablation study to highlight the benefits of training \MethodName{} with a larger and more diverse data mixture while keeping the model architecture fixed. We train versions of \MethodName{}-edge, each on a single modality, and compare them with our multi-modal \MethodName{}-edge policy. Table~\ref{tab:ablation} summarizes the datasets used for training model and reports the goal-reaching rates for each task. For this evaluation, we also introduce satellite images as a goal modality and assess performance under the same environments and setup as 2D goal-pose navigation. As shown in Table~\ref{tab:ablation}, \MethodName{} demonstrates clear advantages, particularly in language- and satellite image-conditioned navigation. By jointly training across multiple modalities, \MethodName{} learns core navigation behaviors from all datasets and achieves stronger generalization to unseen environments compared to single-modality policies trained on limited data. In addition, we observe that including highly varied cross-embodiment data, the BDD-V dataset, enables the policy to succeed in roughly half of the failed cases of a policy trained without BDD-V, highlighting the ability of \MethodName{} to ingest and benefit from diverse data sources.
%
%%SL.9.7: This seems like an interesting result, but I didn't fully understand the explanation here. What is the difference between our method and the baseline here? Is it just the inclusion of the other tasks?
%%NH.9.8: I tried to fix.
%
\subsection{Omni-modal conditioning with \MethodName{}}
By training on omni-modal task representations, \MethodName{} can learn to follow multiple goal signals.
Towards answering {\bf Q2}, we conduct experiments where tasks are specified by providing both 2D goal poses (\emph{where?}) and behavioral language instructions (\emph{how?}) in 10 different environments. In each environment, we provide a 2D position located 25–100 meters from the robot’s initial position, along with a language prompt that specifies a behavior that the robot must follow to reach the goal, such as “move along the wall”, “move on the grass”, and “move between objects A and B”. These prompts are out-of-distribution (OOD) and not included in the training dataset. To our knowledge, no existing work has explored such a challenging composition of modalities in navigation.

Since no existing methods can handle both modalities simultaneously, we compare our approach to \emph{specialist} baselines trained with 2D pose conditioning (NoMaD and MBRA-pose, Table~\ref{tab:flex}). \MethodName{} demonstrates the ability to attend to the information of both the prompt and the goal pose, achieving 80\% success and experiencing only a 5\% drop in language capabilities from Table~\ref{tab:individual}. The smaller \MethodName{} variant fails to handle the language instructions due to limited modal capacity. The other VLA baselines (SmolVLA and MiniVLA) are also unable to solve the task.

\begin{table*}[t]
\vspace{2mm}
\begin{minipage}{0.28\textwidth}
\centering
    \vspace{1mm}
    \resizebox{1.0\columnwidth}{!}{
  \begin{tabular}{lcccc} \toprule 
    \multicolumn{1}{l}{Method} & SR & Prog. & Behavior \\ \midrule
    SmolVLA~\cite{shukor2025smolvla} & 0.10 & 0.35 & 0.20 \\    
    MiniVLA~\cite{belkhale2024minivla} & 0.30 & 0.53 & 0.00 \\
    NoMaD~\cite{sridhar2024nomad}  & 0.40 & 0.57 & -- \\
    MBRA-pose~\cite{hirose2025learning} & 0.70 & 0.80 & -- \\   
    \ours \MethodName{}~\cite{kim2025fine} & \bf{0.80} & \bf{0.86} & \bf{0.60} \\    
    \ours \MethodName{}-edge~\cite{shah2023vint} & 0.30 & 0.56 & 0.10 \\    
    \bottomrule       
  \end{tabular}%
    }
    \captionof{table}{\MethodName{} excels at omni-modal navigation, reaching a goal position (\emph{where}) while following language instructions (\emph{how}).}
    %\captionof{table}{\MethodName{} excels at omni-modal navigation task of reaching a goal position (\emph{where}) while following language instructions specifying behavior (\emph{how}).}
    \label{tab:flex}
\end{minipage}%
\hspace{0.01\textwidth}
\begin{minipage}{0.28\textwidth}
\centering
  \vspace{1mm}
  \resizebox{1.0\columnwidth}{!}{
  \begin{tabular}{lccc} \toprule 
    Modality & Fine-tune & SR & Prog.  \\ \midrule
    Sat. goal image & & 0.57 & 0.75 \\    
    Sat. goal image & \checkmark & 0.83 & 0.92 \\  
    2D goal pose & & 0.81 & 0.90 \\    
    2D goal pose & \checkmark & 0.86 & 0.96 \\  
    \bottomrule       
  \end{tabular}%
  }
\captionof{table}{{\bf Adapting \MethodName{} -edge to new environments.} We adapt our model using only 1.2 hours of data, focusing on satellite goal images and 2D goal poses.}
  \label{tab:capablity_data} 
\end{minipage}
\hspace{0.01\textwidth}
\begin{minipage}{0.4\textwidth}
  \centering
  \includegraphics[width=0.99\hsize]{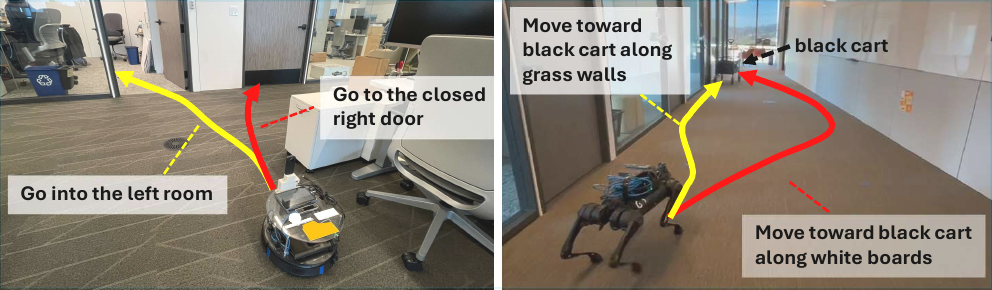}
    \captionof{figure}{\small {\bf Deploying \MethodName{} on multiple embodiments.} We deploy our policy on the Vizbot and Unitree Go1 robots. Our policy can follow natural language instructions out of the box and reach the targets.}
  \label{f:cross}
\end{minipage}
\vspace*{-2.2em}
\end{table*}

\subsection{Adapting \MethodName{} to new goal modalities}
To evaluate \MethodName{}'s capabilities as a ``foundation model'' for navigation and answer {\bf Q3}, we assess three aspects: (i) learning a new goal modality, (ii) fine-tuning on new datasets, and (iii) controlling new robot embodiments.
%we evaluate three aspects of the model: its ability to (i) learn a new goal modality, (ii) fine-tune on new datasets, and (iii) control new robot embodiments. 
%For (i), we setup a controlled experiment where we pre-train \MethodName{}-edge \emph{without} the satellite modality. We then freeze a majority of the model and swap the vision encoders for the egocentric image goals with one for satellite images. 
For (i), we set up a controlled experiment by pre-training \MethodName{}-edge \emph{without} the satellite modality. We then freeze most of the model and replace the vision encoders for egocentric image goals with one for satellite images, training only the swapped encoder.
With this model, we evaluate the adaptability of our model to leverage new goal modalities. We find that \MethodName{}-edge can indeed adapt to this new modality, improving significantly over a specialist policy trained only for satellite goal navigation (0.19 $\rightarrow$ 0.62). This suggests that the cross-modal representations learned by \MethodName{} can facilitate learning new tasks, while retaining useful navigation affordances learned during pre-training.
%%NH.9.15 I think CG misunderstand that Table VI is the results for the above evaluation. But is is wrong.
%%CG.9.12 I think this needs to be clarified. In the table, there is also 2D goal pose, but this was not mentioned originally - also when you fine-tune, I assume this means that you fine-tune the whole model and when it is not fine-tuned its just the visual encoder?
%%SL.9.7: it's unclear to me how this experiment is different from the vint-based satellite image experiment from before (could it make sense to merge these, and cover the vint satellite stuff here instead of earlier?)
%%NH.9.9: Hmm.... I tried to fix! Since we are using many spaces to show just one bottom line in Table V, I am thinking that we do not need to shown Table V and explain in the text to exclude the redundant.

To evaluate (ii), we conduct two experiments to investigate \MethodName{}'s capability to adapt to new data: (a) adapting to new environments using a small dataset, and (b) learning a new language domain from a new available dataset. For (a), we collect 1.2 hours of data, including 2D goal poses and satellite images, from test environments unseen during training, and fine-tune the trained \MethodName{}-edge to evaluate adaptability. Table~\ref{tab:capablity_data} shows that the fine-tuned policy improves performance on both modalities, suggesting that \MethodName{} can effectively adapt to new environments.
% I'm kind of confused by this section, I think you have to be very explicit about why these environments are very different, and you do not describe or give images visualizing what this means. 
For (b), we fine-tune our \MethodName{} using the recently published CounterfactualVLA dataset, which contains more diverse language prompts (see \cite{glossop2025cast}). To stabilize fine-tuning, we sample batches to maintain the balance between modalities used in our prior training. Specifically, half of the data previously sampled from LeLaN is replaced with CounterfactualVLA data when fine-tuning \MethodName{}. 
%The fine-tuned policy improves performance across the board (Table~\ref{tab:individual}), 
The fine-tuned policy achieves scores (SR, Behavior, SR$^S$, SR$^C$) = (0.825, 0.700, 1.000, 0.765), improving over the original scores (0.725, 0.650, 1.000, 0.647), 
demonstrating the ability of \MethodName{} to learn new language domains from emerging datasets. Because the CounterfactualVLA dataset provides both raw and counterfactual robot actions corresponding to diverse language prompts, our policy learns better collision avoidance and instruction-following, with notable improvements in “Behavior” and “SR$^C$”. 
%
%NH.9.11 : We need more reasonable reasons to explain the gap.
% I also think this section could be condensed into one section. It could basically just show the results on adding more data for different modalities and showing the marked improvement with (rather than having it as part a and part b) 

To show the generality of our model trained on data from several robots for (iii), we demonstrate deployment on two additional platforms: the wheeled mobile robot VizBot and the quadruped robot Go1. We mount different cameras on the robots and test them on the most challenging language-conditioned navigation tasks. As shown in Fig.~\ref{f:cross}, our best \MethodName{} enables both robots to follow language instructions and reach the target locations. 
%Detailed robotic behaviors are provided in the supplemental videos.

\section{CONCLUSIONS}
In this work, we introduced \MethodName{}, an omni-modal vision-language-action model for robot navigation. Our policy flexibly interprets multiple goal modalities, including language prompts, 2D poses, egocentric goal images, and their combinations, within a unified framework. By initializing from pre-trained VLAs, \MethodName{} leverages Internet-scale knowledge while being trained on over 9,500 hours of robotic navigation experience across ten robot platforms.

Extensive real-world evaluations show that \MethodName{} consistently outperforms prior baselines, achieving stronger performance across modalities, robust out-of-distribution generalization, and the ability to follow diverse multi-modal tasks. Furthermore, we demonstrate foundation model qualities, including adaptability to new modalities (e.g., satellite imagery), efficient fine-tuning with extra data, and cross-embodiment transfer across different robot platforms.
%Extensive real-world evaluations show that \MethodName{} consistently outperforms prior baselines, achieving stronger performance across modalities, robust out-of-distribution generalization, and the ability to follow diverse multi-modal task specifications. Furthermore, we demonstrated foundation model qualities, including adaptability to new modalities (e.g., satellite imagery), efficient fine-tuning with additional data, and cross-embodiment transfer across different robot platforms.

We believe \MethodName{} represents a significant step toward broadly generalizable navigation policies. To further advance this direction, we will release our models and training code as a resource for developing scalable and flexible robot navigation models. In particular, there remains room for improvement in language-conditioned navigation, both in goal-reaching and instruction-following. Progress in this area will benefit greatly from larger and more carefully curated datasets, which we hope the community will pursue.

\section*{Acknowledgments}
This research was supported by Berkeley AI Research at the University of California, Berkeley and Toyota Motor North America. 
And, this work was partially support by ARL DCIST CRA W911NF-17-2-0181 and the NSF under IIS-2150826.
%And, this work was partially supported by DARPA TIAMAT, ARL DCIST CRA W911NF-17-2-0181, NSF IIS-2246811, and NSF IIS-2150826.
We thank Kyle Stachowicz and Satoshi Koide for discussing the model architecture and its implementation to train our \MethodName{}.
And we thank Frodobots AI for providing robot hardware for our evaluations.


%%%%%%%%%%%%%%%%%%%%%%%%%%%%%%%%%%%%%%%%%%%%%%%%%%%%%%%%%%%%%%%%%%%%%%%%%%%%%%%%
%%%%%%%%%%%%%%%%%%%%%%%%%%%%%%%%%%%%%%%%%%%%%%%%%%%%%%%%%%%%%%%%%%%%%%%%%%%%%%%%
\bibliographystyle{IEEEtran}
\vskip-\parskip
\balance
\begingroup
\footnotesize
%%\scriptsize
%\balance
%%\bibliography{egbib_full.bib}
\bibliography{egbib_old}
\endgroup
%\vfill
\section*{APPENDIX}
%\appendix
%
\subsection{\MethodName{}-edge based on vision-based navigation policies}
\label{app:vint}
\begin{figure}[h]
  %\vspace{2mm}
  \begin{center}
      \includegraphics[width=1.0\hsize]{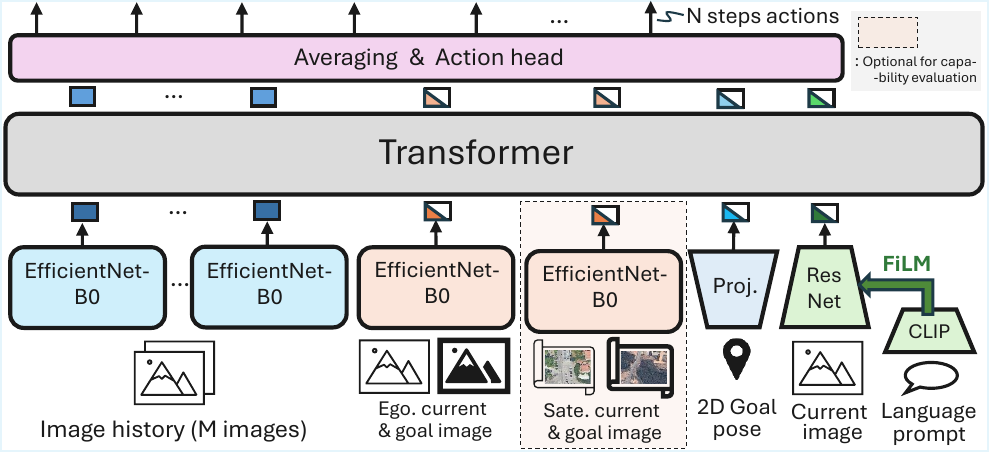}
  \end{center}
    %\vspace{-3mm}
	\caption{\small {\bf Network architecture of \MethodName{}-edge based on the vision-based navigation policies.}}
  \label{f:network_vint}
  %\vspace*{-1.0em}
\end{figure}

In addition to our VLA-based architecture, we design another network, \MethodName{}-edge (Fig.\ref{f:network_vint}), based on prior single-modality vision-based navigation policies such as ViNT\cite{shah2023vint}, NoMaD~\cite{sridhar2024nomad}, MBRA~\cite{hirose2025learning}, and LeLaN~\cite{hirose2024lelan}. Building on ViNT for egocentric goal image-conditioned navigation, we add a projector for 2D goal-pose conditioning and ResNet and CLIP networks with FiLM for language prompt conditioning~\cite{hirose2024lelan,doshi2024scaling}. Similar to the VLA-based model, we design an attention mask according to the selected modality $t_m$ during data sampling. Unlike the VLA-based network, this model employs early fusion, conditioning tokens on each modality before feeding them into the transformer. To maintain temporal consistency with a lightweight architecture, we feed tokens from the last $M=5$ image steps. Following NoMaD~\cite{sridhar2024nomad}, we compute the mean of the transformer-generated tokens and pass them to the action head to produce actions $\{\hat{a}_i\}_{i=1\ldots N}$. Pre-trained weights are used for the EfficientNet-B0, ResNet, and CLIP elements as shown in Fig.~\ref{f:network_vint}.
\subsection{Reannotation model for BDD-V dataset}
\label{app:bdd_annot}

BDD-V~\cite{xu2017end} consists of observations captured by a smartphone mounted on a car dashboard, paired with GPS signals used as actions in the original paper. Compared to other datasets, BDD-V is larger and covers more diverse environments. However, directly using GPS-based actions for training is challenging: they are imprecise due to GPS uncertainty, collected at 1 Hz (vs. 3 Hz for other datasets), and the vehicle speed is about 40 times faster than the mobile robots used in other datasets (0.5 m/s).

To address this embodiment gap, we adapt the reannotation approach in \cite{hirose2025learning} and train a version of the MBRA model to generate feasible trajectories for our target robots. The original MBRA penalizes virtual collisions using estimated 3D points, but BDD-V images often include the dashboard, whose appearance varies with camera pose and can cause false collisions, trapping the model. We resolve this by training MBRA on a combination of GNM and BDD-V, applying the collision-avoidance objective only to GNM. This allows the model to implicitly learn collision avoidance from GNM while adapting to BDD-V’s visual distribution. Additionally, we constrain linear and angular velocities to 0.0–0.5 m/s and ±1.0 rad/s, and enforce the kinematic model of a coaxial two-axle robot, ensuring generated trajectories are consistent with other datasets.
\subsection{Pre-trained checkpoints and backbones for the VLAs}
\label{app:pretrained_vla}
Table~\ref{tab:arch_ablation} lists the language and visual backbones for versions of \MethodName{} and several VLA baselines. These models span a range of sizes and architectures, allowing us to evaluate how pre-trained models contribute prior knowledge to learning \MethodName{}. Further details on the pre-trained checkpoints and backbones are in the original papers.
\begin{table}[ht]
  %\vspace{0mm}
  \begin{center}
  %\vspace{-2mm}
  \resizebox{0.99\columnwidth}{!}{
  \begin{tabular}{lcccc} \toprule 
    Model name & Language backbone & Visual backbone & Model size \\ \midrule
    SmolVLA~\cite{shukor2025smolvla} & SmolLM2 & SigLIP & 500M \\    
    MiniVLA~\cite{belkhale2024minivla} & Qwen2.5-0.5B & DINOv2+SigLIP & 1.0B \\   
    CounterfactualVLA~\cite{glossop2025cast} & Gemma-2B & SigLIP & 2.9B \\
    \midrule
    \ours \MethodName{}-edge & CLIP & EfficientNet-B0 & 50M \\    
    \ours \MethodName{} & Llama2-7B & DINOv2+SigLIP & 7.5B \\
    \bottomrule       
  \end{tabular}%
  }
  \end{center}
  %\vspace{-2mm}
  %\caption{We compare the two variants of \MethodName{} with several VLA baselines: the Gemma-based CounterfactualVLA, as well as omni-modal conditioning recipe applied to SmolVLA and MiniVLA.}
  \caption{We compare two variants of \MethodName{} with VLA baselines: the Gemma-based CounterfactualVLA and the omni-modal conditioning recipe applied to SmolVLA and MiniVLA.}
  %\vspace{-3mm}
  \label{tab:arch_ablation}
\end{table}
\subsection{Breakdown of GNM and LeLaN dataset mixture}
\label{app:gnm_lelan}
The GNM mixture in Table~\ref{tab:dataset} comprises seven publicly available datasets—RECON~\cite{shah2021rapid}, CoryHall~\cite{kahn2018self}, TartanDrive~\cite{triest2022tartandrive}, Seattle~\cite{shaban2022semantic}, SCAND~\cite{karnan2022socially}, SACSoN~\cite{hirose2023sacson}, and GO Stanford 4~\cite{hirose2019deep}—combined as in \cite{shah2023gnm}. Using the curated expert actions as both action commands and 2D goal poses, \MethodName{} learns policies conditioned on egocentric images, 2D poses, or their combinations.

LeLaN~\cite{hirose2024lelan} provides synthetic action labels, language prompts, and object poses from SACSoN, GO Stanford 2 $\&$ 4~\cite{hirose2018gonet,hirose2019deep}, HumanWalk, and YouTube videos. We leverage these annotations to train \MethodName{} for language grounding.
\subsection{Objective design}
\label{app:obj}
Following \cite{hirose2024lelan}, we introduce the additional objectives $J_{obj}$ and $J_{sm}$ in addition to the main objective $J_{il}$ and train our \MethodName{} policy, $\pi_\theta$ to minimize the entire $J$. 
%
%%SL.9.6: I find this equation and the paragraph that follows very confusing. Are we actually using multiple objectives, or are we just constructing labels differently for different datasets? If multiple objectives, then we should find a way to explain this better, if different labels for different datasets, then we should present this differently, instead talking about how the data (and labels) are constructed rather than presenting it as multi-objective.
%%NH.9.8: In Slack, I said that I can move J_obj and J_sm into the implementations or appendix. But I think that it is also annoying for readers. So I tried to accurately explain the objective one by one.
\begin{equation}
    \min_\theta \,J := J_{il} + m_{obj}J_{obj} + J_{sm},  
    \label{eq:objective}
\end{equation}
$J_{obj}$ is designed as $(p_{obj} - \hat{a}_N)^2$ to encourage the policy to generate actions that move toward the target object pose $p_{obj}$ in language-conditioned navigation.
%%SL.9.13: if this is only for LeLaN, can we directly say that? else a bit confusing (it's implied later, but we could say directly: all datasets have the standard imitation learning objective J_im. In addition, for LeLaN, we follow [11] and add J_obs, the object position reaching objective given by [whatever]. Finally, we regularize the model to provide for smooth trajectories by adding a regularizer J_sm, which encourages sequential actions to be similar, given by $$
Following \cite{hirose2024lelan}, we penalize the $N$-th action $\hat{a}_N$ to be close to $p_{obj}$. Since $J_{obj}$ is only for the LeLaN dataset to learn language grounding, we set $m_{obj}=1$ for the LeLaN dataset, otherwise $m_{obj}=0$ to mask out $J_{obj}$. In addition, $J_{sm} = \frac{1}{N-1}\sum_{i=1}^{N-1}(\hat{a}_{i+1}- \hat{a}_i)^2$ is the objective to minimize the action deltas for regularization. 
\end{document}